\documentclass[sn-apa]{sn-jnl}
\jyear{2022}
\usepackage{microtype}

\begin{document}

\title[{UstanceBR}: a social media language resource for stance prediction]{{UstanceBR}: a social media language resource for stance prediction}

\author[1]{\fnm{Camila} \sur{Pereira}}\email{camilafpp@usp.br}
\author[1]{\fnm{Matheus} \sur{Pavan}}\email{matheus.pavan@usp.br}
\author[1]{\fnm{Sungwon} \sur{Yoon}}\email{sungwon.yoon@usp.br}
\author[1]{\fnm{Ricelli} \sur{Ramos}}\email{ricellimsilva@gmail.com}
\author[1]{\fnm{Pablo} \sur{Costa}}\email{pablo.costa@usp.br}
\author[1]{\fnm{La\'is} \sur{Cavalheiro}}\email{lalas.teste@gmail.com}
\author*[1]{\fnm{Ivandr\'e} \sur{Paraboni}}\email{ivandre@usp.br}

\affil*[1]{\orgdiv{School of Arts, Sciences and Humanities}, \orgname{University of S\~ao Paulo}, \orgaddress{\street{Av Arlindo Bettio 1000}, \city{S\~ao Paulo}, \postcode{03828000}, \state{SP}, \country{Brazil}}}

\abstract{
This work introduces \mbox{UstanceBR}, a novel social media corpus in the Brazilian Portuguese Twitter/X domain for target-based stance prediction. The corpus comprises 86.8 k labelled stances towards selected target topics, and extensive network information about the users who published these stances on social media. In this article we describe the corpus data, and a number of usage examples in both in-domain and cross-target stance prediction based on text- and network-related information, which are intended to provide initial baseline results for future studies in the field.}

\keywords{Corpora, stance prediction, social media, Portuguese NLP}

\maketitle

\section{Introduction}
\label{sec.intro}

Stance prediction \citep{st-survey2,st-survey1} is the  task of automatically determining for/against attitudes  towards a given target (e.g., a politician, climate change, etc.) from text and, in some cases, from multimodal data \citep{corp-islands,corp-homophily,corp-radical}. The task is analogous to sentiment analysis, but distinct in that sentiment and stance do not necessarily correlate \citep{st-survey2}. Thus, for instance, an attitude in favour of the target `World Cup' may express a negative sentiment, as in `{\it I hate having to work on match days}'.

Computational models of stance prediction often rely on supervised machine learning methods based on annotated corpora, often in in-domain \citep{semeval2016} or more recently in zero-shot and cross-target settings \citep{corp-zeroshot}. Accordingly, purpose-built corpora have become increasingly common since the \mbox{SemEval-2016} English shared task in \cite{semeval2016}, and are now available in multiple languages and domains \citep{corp-iberaval2017,corp-sardine,corp-arabic,corp-multicovid}. 

Most of the existing resources focus on a well-defined set of target entities (often of a political or moral nature, as in \cite{semeval2016,bracis-pavan}), in so-called target-based stance prediction, but others have addressed the issue of stance towards propositions in quoted text \citep{corp-facebook,corp-arabic}. Our own  work focuses on the former, that is, on the development of a novel language resource for target-based stance prediction, and which  is intended to fill a number of  gaps in the field. First of all, we notice that our target language - Brazilian Portuguese, which is spoken by some 214 million individuals - still lacks resources for target-based stance prediction. This represents a resource gap  not only at the language level, but also at the domain level since many of the target topics addressed in existing work (e.g., the Catalonia independence in \cite{corp-iberaval2017}, Danish immigration policies in \cite{corp-danish}, etc.) would be unfamiliar among Portuguese speakers.


We notice also that most of the existing resources usually provide text-level labelling (e.g., individual tweet instances may be labelled as being for/against a given target, etc.), and that few (e.g., \cite{corp-ok}) have addressed the  issue of user-level stance prediction \citep{jiis-arthur} by providing user- or timeline-level stance labels. Although instance- and user-level prediction  arguably represent distinct task definitions, we notice that the ubiquity  of social media data allows us to build a collection of individually-labelled instances accompanied by additional user-generated contents - including their full public timelines, which would support both kinds of investigation at once. 

Finally, as pointed out in \cite{st-survey2} and others, stance prediction does not need to rely solely on text data and, at least when applied to social media settings, models of this kind may greatly benefit from the well-known principle of  homophily, that is, the notion that connection is motivated by similarity. This suggests that purely text-based  resources may be enriched with a wide range of non-textual information (e.g., friends connections, interactions, etc.) as indeed a number of studies have recently considered (e.g., \cite{corp-islands,corp-homophily,corp-sardine}.) 


Based on these observations, we envisage creating a novel social media corpus in the Brazilian Portuguese Twitter/X domain,  hereby called \mbox{UstanceBR}. The corpus follows the \mbox{SemEval-2016} tradition of using a pre-defined set of targets \citep{semeval2016}, but pays regard to both explicit and implicit stance as in \cite{corp-deinstance}  with a focus on heavily polarised topic pairs as in \cite{corp-hate}. 

Although not the first Twitter/X stance corpus for Portuguese \citep{ranlp-wesley}, \mbox{UstanceBR} is, to the best of our knowledge, the first resource of its kind to include extensive network information covering not only the individuals who produce every stance, but also their entire social media friends and followers lists\footnote{Some limited information of this kind has been made available for other languages in \cite{corp-sardine} and \cite{corp-lai-muti2020}.}. This may potentially support, e.g., future studies on the role of polarisation in stance prediction,  which are not necessarily limited to the use of text data. 

In addition to describing the corpus itself, in this article we  report a number of usage examples in both in-domain and cross-target stance prediction from \mbox{UstanceBR} non-textual data. The main intended contributions are as follows.

\begin{itemize}
\item{Large corpus of explicit and implicit stances towards polarised targets in the Portuguese Twitter/X domain.}
\item{Accompanying timeline text data and network information for both user-level  stance prediction based on non-textual data.}
\item{Reference results for in-domain stance prediction based on text and network data alike, and cross-target stance prediction from text.}
\end{itemize}

\section{Background}
\label{sec.background}

Table \ref{tab.corpora} shows a number of selected  studies that have recently produced labelled data for stance prediction, categorised according to target type (a pre-defined set of entities  or quoted text, also known as claim-based stance prediction), domain (e=essays, f=facebook, n=news, o=opinions, p=political discourse, t=Twitter/X, r=reddit, we=weibo, w=wikipedia, y=youtube), target language (Ar=Arabic, Ca=Catalan, Ch=Chinese, Da=Danish, De=German, Du=Dutch, En=English, Fr=French, It=Italian, Pt=Portuguese, Sp=Spanish, Tu=Turkish, *=others), the availability of non-text data (net=network features, dem=demographics, or dom=domain information such as political affiliation, etc.), and corpus size in number of instances. We notice, however, that the list is by no means exhaustive, and for further details on resources of this kind we report to \cite{st-survey3}.

\footnotesize{
\begin{table}[ht]
\centering
\caption{\label{tab.corpora}Selected stance prediction corpora.}
\setlength{\tabcolsep}{2pt}
\begin{tabular}{p{0.5\columnwidth} c c l l r}
\hline
Corpus                  & Stance& Domain    & Language   & Non-text & Size\\
\hline
\cite{semeval2016}      & set   & t     & En             &             & 4,163 \\
\cite{corp-essays}      & quote & e     & En             &             & 826 \\
\cite{corp-iberaval2017}& set   & t     & Ca,Sp          & dem         & 10,800 \\
\cite{corp-islands}     & set   & t     & Ar             & net         & 33,024 \\
\cite{corp-multitarget} & set   & t     & En             &             & 4,455 \\
\cite{corp-facebook}    & quote & f     & Ch             & net         & 12,828 \\
\cite{corp-danish}      & set   & p     & Da             & dem,dom     & 898 \\
\cite{corp-homophily}   & set   & t     & It             & net         & 968 \\
\cite{corp-will}        & set   & t     & En             &             & 51,284 \\
\cite{brmoral}          & set   & e     & Pt             & dem         & 4,080 \\  
\cite{corp-french}      & quote & t     & Fr             &             & 5,803 \\
\cite{corp-xstance}     & quote & p     & Fr,It,De       &             & 67,281 \\
\cite{corp-contemporary}& quote & o     & En             &             & 6,094 \\
\cite{corp-cloroq}      & set   & t     & En             &             & 14,374 \\ 
\cite{corp-sardine}     & set   & t     & It             & net         & 3,242 \\  
\cite{corp-zeroshot}    & set   & o     & En             &             & 23,525 \\
\cite{corp-lai-muti2020}& set   & t     & En,Fr,*        & net,dom     & 14,440 \\
\cite{corp-brexit}      & set   & t     & En             & net,dom     & 1,760 \\
\cite{corp-arabic}      & quote & n     & Ar             &             & 4,063 \\
\cite{corp-hate}        & set   & t     & En             &             & 3,000 \\
\cite{corp-pstance}     & set   & t     & En             &             & 21,574 \\
\cite{corp-exasc}       & set   & t     & Ar             &             & 9,566 \\
\cite{corp-deinstance}  & quote & p     & De             &             & 1,000 \\
\cite{corp-portuguese}  & quote & t     & Pt             &             & 16,655 \\
\cite{corp-ok}          & set   & r     & En             & net         & 2,716,998 \\ 
\cite{corp-radical}     & set   & r     & En             & net         & 3,526\\
\cite{corp-multicovid}  & set   & t     & Fr,De,*        &            & 17,934 \\
\cite{mawqif}           & set   & t     & Ar             &             & 4,121 \\
\cite{turkish}          & set   & t     & Tu             &             & 4,502\\
\cite{c-stance}         & set   & we    & Ch             &             & 48,126\\
\cite{orchid}           & quote & w     & Ch             &             & 14,133 \\
\cite{ez-stance}        & quote & t     & En             &             & 47,316\\
\cite{charfi2024}       & set   & t,y   & Ar             &             & 5,657\\
\hline
\end{tabular}
\end{table}
}
\normalsize

Existing corpora for stance prediction fall into two broad categories according to the kind of target under consideration, hereby identified as {\em set} and {\em quote} stance definitions. In the more standard, \mbox{SemEval}-like task definition, a pre-defined set of targets is considered. These may consist of a single subject, as  the Catalonia independence movement in the IberEval-2017 corpus \citep{corp-iberaval2017}, a small and well-defined set of topics (often following some of the \mbox{SemEval} topics of political or moral nature, such as Trump, Hillary, abortion legislation,  etc.), or ultimately any topic covered in the corpus regardless of their actual number of instances (e.g., \cite{corp-zeroshot,corp-exasc}). 

In quote- or claim-based stance prediction, by contrast, the intended target consists of a quoted piece of text (e.g., a previous tweet or idea under discussion, a proposition, a question, etc.) In these cases, the notion of a well-defined target may be secondary or even absent from the data, and different instances may be simply regarded as referring to different targets. As a result, the number of possible targets is usually large \citep{corp-french,corp-xstance,corp-arabic,corp-deinstance,corp-portuguese}, and although the goals remain essentially the same as in target-based stance prediction, modelling the intended target as a piece of quoted text may suggest a different computational approach based on, e.g., text similarity between target and stance texts.   

Regarding the text genre adopted by existing resources, we notice that the use of data in the Twitter/X domain (and to a lesser extent other social media) is common and, as expected, there is a significant presence of corpora in the  English language. Most resources convey purely text data, with a number of exceptions including mainly network-related information such as interaction features (e.g., replies, retweets, discussion contexts, etc.) or user demographics,  as in \cite{corp-iberaval2017,corp-danish}. 

Most recent resources are considerably larger than the original \mbox{SemEval} corpus, but the largest  are  created either by using label propagation \citep{corp-islam}, by using existing labels available from the data such as response scores \citep{corp-xstance}, or are simply labelled at the user level rather than at the instance level \citep{corp-ok}. 

Finally, we notice that most existing corpora use a three- or four-class label definition (e.g., `for' and `against', accompanied by `neutral', `neither', or both). An exception is the corpus in \cite{corp-xstance}, which focuses on the for/against extremes only, and which has been partially extended in \cite{corp-deinstance} to include more fine-grained labels representing either explicit or implicit stance information as in \cite{corp-radical}.


\section{UstanceBR}
\label{sec.current}

The existing work described in the previous section suggests that our target language - Brazilian Portuguese - remains  underrepresented in the field, the exceptions being the quote-based corpus of European Portuguese tweets in \cite{corp-portuguese}, and Brazilian Portuguese essays corpus BRmoral \citep{brmoral}. As a result,  target topics covered by previous work do not generally match our local interests (e.g., Brazilian politics). An  exception is the corpus of stances towards the use of Hydroxychloroquine as a bogus Covid-19 treatment in \cite{corp-cloroq}, available in the English language. None of these  resources, however, includes non-textual data.  

Based on these observations we envisage creating a novel social media corpus of labelled tweets and accompanying network information in the Brazilian Portuguese language. The corpus, called \mbox{UstanceBR},  focuses on six targets organised in polarised pairs (two Brazilian presidents, two kinds of medication, and two institutions) mostly favoured by either conservative or liberal individuals. The  target pairs consist of two politicians of left/right leaning, two medications that were widely debated during the Covid-19 pandemic (a vaccine generally favoured by liberals, and a drug promoted by conservatives as alternative treatment), and two institutions (a TV channel often attacked by the conservatives, and the church). 

The corpus data is organised around three related datasets: a {\em stance} corpus conveying 86.8 k tweets (1.7 million words) labelled as for/against/other towards the six targets under consideration, a {\em timelines} corpus conveying 34.8 million public tweets (511.8 million words) published by every individual who authored a for/against stance, and additional {\em network} data conveying (4.5 million unique friend and follower identifiers, and additional user mentions.)

\subsection{Data collection and annotation}

Data collection started by searching publicly available Portuguese-speaking Twitter/X data for keywords representing the six targets of interest. These were the Brazilian politicians `Lula' and `Bolsonaro', `Cloroquina' (Hydroxychloroquine), `Coronavac' (Sinovac vaccine), `Globo' (a popular TV channel), and `igreja' (church). About 30k matching tweets from the 2018-2020 period were retrieved for each target. 

Duplicates were removed  from the raw data, and each target was annotated by two  independent judges over a six-month period. As a means to focus on the more clear cases of for/against stances, we decided to keep in the final dataset only those instances for which there was agreement between the two judges. As for the `other' category, however, we kept all instances that were marked as `other' by  at least one of the two annotators, the underlying assumption being that these cases were not sufficiently clear to qualify as a proper stance towards the target.  

Annotation  was carried out until each annotator  obtained at least 4000 positive (for) and further 4000 negative (against) instances for each target regardless of the number of `other' instances that they came across during the process. For most classes, however, the final corpus size turns out to be below 4000 instances as we only kept the instances in which there was total agreement between judges.  

Annotation followed a protocol similar to \cite{semeval2016} and others, in which judges were requested to decide whether a given tweet expressed a stance towards the target or not and, if so, whether the stance would be in favour or against it. Thus, all tweets mentioning a keyword with a different meaning, or different context, were marked as `others'. These include, for instance, the use of homonyms (e.g., the word `church' referring to a physical building rather than a religious institution), the expression of factual information (e.g., `{\em Lula was the president at that time}') and others. 


One significant difference between the present protocol and those followed in \cite{semeval2016} and similar studies is the definition of what actually counts as a stance. More specifically, unlike  annotation tasks in which linguistic topic may coincide with the stance target, annotators were instructed to take a broad perspective and consider how the tweet reflected upon the target even if the target was not the main topic under discussion. Thus, for instance, a corruption scandal in the government of Lula or Bolsonaro was to be interpreted as a stance against them even if the main topic under discussion was the scandal, and not the person of the president himself. As a result, \mbox{UstanceBR} may be seen as a collection of more explicit (e.g., `{\em TV Globo has done a great job}') and less explicit (`{\em I love the film that TV Globo is broadcasting right now}') stances towards the selected targets. Similar distinctions have been addressed in \cite{corp-deinstance}. 

Table \ref{tab.kappa} presents average agreement and Cohen's Kappa results for the six target topics. 

\begin{table}[ht]
\caption{\label{tab.kappa}Annotation agreement.}
\centering
\begin{tabular}{ l c c  }
\hline
Target	     & \% Agreement & Cohen's Kappa \\
\hline
Bolsonaro    &  89.9 & 0.79 \\
Lula 	     &  57.7 & 0.24 \\
Hydrox. 	 &  91.7 & 0.82 \\
Sinovac 	 &  89.2 & 0.78 \\
Globo TV 	 &  88.2 & 0.76 \\
Church 	     &  63.3 & 0.28 \\
\hline
\end{tabular}
\end{table}

From these results we notice that inter-annotator agreement  was very high or substantial for four targets  (Hydroxychloroquine, Bolsonaro, Sinovac and Globo TV), and fair for the other two  (Lula and Church). Upon manual inspection, lower agreement was found to be largely due to a more conservative view on so-called implicit stances (e.g., `{\em The film Globo TV showed last night was great}', which reflects only indirectly upon the TV network), which were often dismissed by one of the annotators as `others' instead of `for' or `against'.


For the set of agreed for/against stances, we also collected the entire public timeline of every user, and their lists of friends and followers. Finally, the stance dataset was expanded to include those tweets marked as `other' as discussed above, bearing in mind that these are not accompanied by user timelines. 

Each of the six subsets in the stance corpus was  subject to a pseudo-random 75-25 train-test split as a means to provide reference results for future work, as reported in Section \ref{sec.indomain}. In previous work \cite{bertabaporu} based on the present corpus, a class-balanced train-test split (hereby called r2) was considered. In the r2 split, texts written by the same individual may appear in either train or test, which is suitable for standard text-based stance prediction but not for network-based methods. For that reason, a second train-test split, hereby called r3, has been  created from the same corpus data. The r3 split ensures that train and test subsets convey text written by disjoint sets of users and, unlike the r2 split, is not class-balanced. However, by guaranteeing that network information used as train data do not appear in the test data, the r3 split is suitable for stance prediction based on network-related features.

Both r2 and r3 splits are available from the corpus download  repository\footnote{\url{https://drive.google.com/drive/folders/1qThfcIe0HjwVbsDVgkot-AqgnXoJdB0K}} as sets of (labelled) tweet identifiers, hence in accordance to  Twitter/X  privacy policies that forbid the reproduction of tweet contents.


\subsection{Descriptive statistics}

\mbox{UstanceBR} main component, the {\em stance corpus} proper, is a collection of labelled tweets in the Portuguese language not unlike the \mbox{SemEval}  corpus in \cite{semeval2016}. Descriptive statistics are summarised in Table \ref{tab.st-stats}.

\begin{table}[ht]
\caption{\label{tab.st-stats}Stance corpus descriptive statistics.}
\centering
\begin{tabular}{ l   c c c c c c }
\hline
Target	     & Against & For  & Other    & All & Words        & W/tweets\\
\hline
Bolsonaro    & 5,565   & 3,849 & 1,668   & 11,082 & 259,521   & 23.4 \\
Lula 	     & 4,514   & 3,806 & 13,934  & 22,254 & 422,064   & 19.0 \\
Hydrox. 	 & 3,978   & 4,017 & 1,785   & 9,780  & 277,824   & 28.4 \\
Sinovac 	 & 4,058   & 3,915 & 932	 & 8,905  & 252,663   & 28.4 \\
Globo TV 	 & 3,341   & 2,672 & 15,253  & 21,266 & 214,876   & 10.1 \\
Church 	     & 3,539   & 3,598 & 6,329   & 13,466 & 322,289   & 23.9 \\
\hline
Overall 	 & 24,995  & 21,857& 39,901  & 86,753 & 1,749,237 & 20.7 \\
\hline
\end{tabular}
\end{table}

We notice that the smaller sizes of some subsets (e.g., Sinovac) are due to the higher proportion of stances (for/against) on social media. In other words, most mentions to Sinovac do represent a stance for/against vaccination, whereas factual mentions to targets such as Lula or Globo TV are more common.  These limitations not withstanding, we notice that the final dataset - comprising 86,753 labelled tweets - is considerably larger than existing work in the field, with the exception of the aforementioned studies that use label propagation (e.g., \cite{corp-islam}) or user-level labelling (e.g., \cite{corp-ok}). 

\mbox{UstanceBR} {\em timelines corpus} is a collection of unlabelled tweets comprising the entire public timeline of the above individuals who provided for/against stances, from which the stances that appeared in the above stance corpus were removed. This dataset is intended to support future studies in user-level stance prediction \citep{corp-ok}. Descriptive statistics are summarised in Table \ref{tab.tl-stats}. 

\setlength{\tabcolsep}{4pt}
\begin{table}[ht]
\caption{\label{tab.tl-stats}Timelines corpus descriptive statistics.}
\centering
\begin{tabular}{ l   c c c c c c c}
\hline
Target	    & Users & Male \%& Tweets(mi) & T/users & Words(mi) & W/tweets & W/users \\
\hline
Bolsonaro 	& 756	& 42.2   & 2,939	   & 3,887.3 & 35,756    & 12.2 	& 47,296.5 \\
Lula 	    & 1,097	& 59.1   & 4,160	   & 3,792.5 & 58,152    & 14.0 	& 53,010.1 \\
Hydrox. 	& 2,298	& 73.9   & 4,904	   & 2,134.1 & 94,845    & 19.3 	& 41,272.7 \\
Sinovac 	& 3,099	& 70.5   & 6,562	   & 2,117.6 & 120,828   & 18.4 	& 38,989.3 \\
Globo TV 	& 1,661	& 41.8   & 6,632	   & 3,992.7 & 81,180    & 12.2 	& 48,874.4 \\
Church 	    & 2,404	& 39.7   & 9,627	   & 4,004.4 & 121,072   & 12.6 	& 50,362.9 \\
\hline
Overall 	& 11,315 & 57.4  & 34,824	   & 3,077.7 & 511,834   & 14.7 	& 45,235.0 \\
\hline
\end{tabular}
\end{table}

Finally, \mbox{UstanceBR} {\em network data} consists of anonymised identifiers representing the lists of friends, followers and user mentions (the latter being computed from the user's timeline texts) of the individuals who provided for/against stances. Descriptive statistics are summarised in Table \ref{tab.net-stats}. 

\begin{table}[ht]
\caption{\label{tab.net-stats}Network data descriptive statistics.}
\centering
\begin{tabular}{ l   c c c }
\hline
Target	    & Friends avg.    & Followers avg. & Mentions avg. \\
\hline
Bolsonaro 	&  774           & 1277          & 106 \\
Lula 	    &  919           & 1687          & 160\\
Hydrox. 	&  1579          & 7665          & 723\\
Sinovac 	&  1588          & 11490         & 598\\
Globo TV 	&  944           & 2270          & 112\\
Church 	    &  931           & 2684          & 119\\
\hline
\end{tabular}
\end{table}

\section{In-domain stance prediction from text and network data }
\label{sec.indomain}


In-domain stance prediction represents the standard task in which both train and test data refer to the same intended target. In what follows we describe initial results for in-domain stance prediction towards each of the six \mbox{UstanceBR} targets independently, and using both text and network data (namely, lists of friends, followers and mentions to other users) as an input. 

For text-based prediction, we use a standard neural architecture with a number of parameters selected through grid search. The input consists  of a token embedding layer provided by BERTabaporu \citep{bertabaporu}, a BERT language model that has been pre-trained on 237 million tweets in Portuguese\footnote{\url{https://huggingface.co/pablocosta/bertabaporu-large-uncased}}.


For network-based prediction, we build three separate models based on friends, followers and user mention lists using the unbalanced r3 train-test corpus split in which train and test users are disjoint. All models follow  a bag-of-users approach based on TF-IDf counts (that is, creating bag-of-friends, bag-of-followers, and bag-of-mentions vectors.) Each of these vectors is taken as an input to a logistic regression classifier using liblinear solver and balanced class weights with optimal penalty (L1 or L2), $C$ (0.1 to 1.9) and tol (0.01 to 1e-05) parameter values obtained through grid search.

Table \ref{tab.indomain} report  results for in-domain stance prediction based on text- and network-related features using the unbalanced r3 train-test corpus split. 

\footnotesize{
\setlength{\tabcolsep}{2pt}
\begin{table}[!htbp]
\begin{center}
\caption{\label{tab.indomain} In-domain stance prediction results based on text- and network-related features using the r3 train-test corpus split. Best F1 scores for each target are  highlighted.}
\begin{tabular}{l | c c c | c c c | c c c | c c c }
\hline
\multicolumn{1}{c|}{}&
\multicolumn{3}{c|}{Text}&
\multicolumn{3}{c|}{Friends}&
\multicolumn{3}{c|}{Followers}&
\multicolumn{3}{c}{Mentions}\\
Target      & P    & R    & F1              & P    & R    & F1     & P    & R    & F1     & P    & R    & F1    \\
\hline
Bolsonaro   & 0.81 & 0.77 & \textbf{0.78}   & 0.75 & 0.75 & 0.75   & 0.89 & 0.65 & 0.70   & 0.72 & 0.73 & 0.73 \\
Lula        & 0.85 & 0.84 & \textbf{0.84}   & 0.73 & 0.73 & 0.73   & 0.72 & 0.71 & 0.70   & 0.71 & 0.71 & 0.71 \\
Hydrox.     & 0.85 & 0.86 & 0.85            & 0.85 & 0.84 & 0.84   & 0.84 & 0.82 & 0.82   & 0.86 & 0.86 & \textbf{0.86} \\
Sinovac     & 0.86 & 0.87 & \textbf{0.86}   & 0.86 & 0.82 & 0.83   & 0.83 & 0.76 & 0.76   & 0.85 & 0.84 & 0.84 \\
Globo TV    & 0.89 & 0.89 & \textbf{0.89}   & 0.62 & 0.59 & 0.59   & 0.62 & 0.57 & 0.56   & 0.65 & 0.61 & 0.60 \\
Church      & 0.88 & 0.89 & \textbf{0.88}   & 0.64 & 0.64 & 0.64   & 0.61 & 0.61 & 0.61   & 0.65 & 0.65 & 0.64 \\
\hline
\end{tabular}
\end{center}
\end{table}
}
\normalsize

Results from Table \ref{tab.indomain} show that text-based models generally outperform the network-based alternatives, even though for some topics the difference is relatively small. This outcome reinforces the view  that social media interaction may play a significant role in stance prediction \citep{corp-islam,corp-islands,corp-facebook,corp-homophily,corp-sardine,corp-lai-muti2020,corp-brexit,corp-ok,corp-radical}.

\section{Cross-target stance prediction from text data}
\label{sec.zeroshot}

From a machine learning perspective, in-domain classification will arguably obtain optimal results as long as suitably labelled  training data is available for every target of interest. However, as the number of possible targets is unlimited, more adaptable approaches are called-for. Among these, a common strategy is the use of cross-target  methods \citep{corp-zeroshot,allaway2022}, that is, building a model from a source domain that happens to be available (i.e., from labelled stances towards some other topic), and then using this model to predict, possibly with some accuracy loss, stances towards an unseen intended target.

An initial investigating of cross-target methods for stance prediction based on a preliminary version of the present corpus data, appeared in \cite{ustancebr1}. Using now the entire dataset, we conducted two sets of pairwise experiments. In the first setting, source data is obtained from the more closely-related target in the corpus, that is, each target is predicted from stances towards its own counterpart (i.e., the president, medication, or institution that is not the target itself.) In the second setting, we attempt to use every possible target other than the target itself as source  data, and report results from the best alternative for each scenario. 

The cross-target experiments use the same in-domain classifiers described in the previous section, varying only the source data. Results from pairwise cross-target stance prediction considering both closely-related target-source domain pairs, and only the best source domain for each target, are shown in Table \ref{tab.zeroshot} using the class-balanced r2 train-test corpus split. 

\setlength{\tabcolsep}{4pt}
\begin{table}[ht]
\begin{center}
\caption{\label{tab.zeroshot} Pairwise cross-target stance prediction results using the r2 train-test corpus split. }
\begin{tabular}{l | l c c c | l c c c}
\hline
\multicolumn{1}{c|}{}& \multicolumn{4}{c|}{Closely-related source domain}& \multicolumn{4}{c}{Best source domain} \\
Target      & Source    & P    & R    & F1     & Source    & P    & R    & F1     \\
\hline
Bolsonaro   & Lula      & 0.59 & 0.57 & 0.56   & Lula       & 0.59 & 0.57 & 0.56  \\
Lula        & Bolsonaro & 0.53 & 0.54 & 0.53   & Sinovac    & 0.73 & 0.72 & 0.70  \\
Hydrox.     & Sinovac   & 0.72 & 0.71 & 0.71   & Sinovac    & 0.72 & 0.71 & 0.71  \\
Sinovac     & Hydrox.   & 0.66 & 0.66 & 0.65   & Lula       & 0.74 & 0.74 & 0.74  \\
Globo TV    & Church    & 0.31 & 0.35 & 0.30   & Lula       & 0.72 & 0.72 & 0.72  \\
 Church      & Globo TV  & 0.31 & 0.34 & 0.31   & Hydrox.    & 0.66 & 0.65 & 0.64  \\
\hline
\end{tabular}
\end{center}
\end{table}

Cross-target stance prediction results are below those obtain by text-based in-domain prediction (left side of Table \ref{tab.indomain}), particularly when using the more closely-related source domain for each target (on the left side of Table \ref{tab.zeroshot}). This suggests that semantic relatedness may not be a major (or at least not the only) factor at play in our current setting, and indeed there seems to be a better source domain for most targets, as shown on the right side of the table. 





\section{Conclusions}
\label{sec.final}

This work has introduced \mbox{UstanceBR}, a novel social media corpus of labelled stance for the Brazilian Portuguese language. The corpus, which comprises  contents produced by over 11 k users of Twitter/X, has been made publicly available for reuse, and it is  suitable for both in-domain and zero-shot/cross-target stance prediction based on text- and network-related data alike.

The initial experiments presently reported, and which are only intended to provide reference results for future work, leave a number of opportunities of investigation beyond standard text-based stance prediction. In particular, we notice that the availability of text- and non-text data may suggest a wide range of, e.g., ensemble methods, combining these knowledge sources. Moreover, we notice also that the availability of users' timelines (in addition to their stance texts) supports the investigation of user-level stance prediction \citep{corp-ok}. These issues are left as suggestions of future work.

\section{Declarations}

\noindent
{\bf Funding}. This work has been supported by FAPESP grant \# 2021/08213-0.\\[1ex]
\noindent
{\bf Authors' contribution}. Conceptualisation, Methodology, Investigation, Writing, Supervision: I. Paraboni. Corpus annotation: C. Pereira, M. Pavan, S. Yoon, R. Ramos, P. Costa. Validation, Formal analysis: C. Pereira. Text-based experiments: M. Pavan. Network-based experiments: L. Cavalheiro. Language models: P. Costa. Manuscript revision: all.\\[1ex]
\noindent
{\bf Conflicts of interest}. The authors declare none.
 
\bibliography{refs}
\end{document}